\documentclass[a4paper,conference]{IEEEtran}
\IEEEoverridecommandlockouts

\ifCLASSINFOpdf
\else
\fi

\usepackage{url}
\usepackage{amssymb,amsmath,bm}
\usepackage{hyperref,color}
\hypersetup{
    colorlinks=true,
    linkcolor=blue,
    urlcolor=blue,
    citecolor=cyan,
}
\usepackage{balance}

\usepackage{graphicx}

\begin{document}
%
\title{Uncertain Label Correction via Auxiliary Action Unit Graphs for Facial Expression Recognition}


\DeclareRobustCommand*{\IEEEauthorrefmark}[1]{%
    \raisebox{0pt}[0pt][0pt]{\textsuperscript{\footnotesize\ensuremath{#1}}}}


%
\author{\IEEEauthorblockN{Yang Liu\IEEEauthorrefmark{1,2}, Xingming Zhang\IEEEauthorrefmark{1}, Janne Kauttonen\IEEEauthorrefmark{3} and Guoying Zhao\IEEEauthorrefmark{2}$^*$}
\IEEEauthorblockA{\IEEEauthorrefmark{1} School of Computer Science and Engineering, South China University of Technology, Guangzhou, China, 510006}
\IEEEauthorblockA{\IEEEauthorrefmark{2} Center for Machine Vision and Signal Analysis, University of Oulu, Oulu, Finland, FI-90014}
\IEEEauthorblockA{\IEEEauthorrefmark{3} Haaga-Helia University of Applied Sciences, Helsinki, Finland, FI-00520}

\thanks{$^*$ Corresponding author}
\thanks{This work was supported by the China Scholarship Council under Grant 202006150091, and the Academy of Finland for Academy Professor project EmotionAI (grants 336116, 345122), and the Ministry of Education and Culture of Finland for AI forum project. The authors wish to acknowledge CSC-IT Center for Science, Finland, for generous computational resources.}
}


\maketitle

\begin{abstract}
  High-quality annotated images are significant to deep facial expression recognition (FER) methods. However, uncertain labels, mostly existing in large-scale public datasets, often mislead the training process. In this paper, we achieve uncertain label correction of facial expressions using auxiliary action unit (AU) graphs, called ULC-AG. Specifically, a weighted regularization module is introduced to highlight valid samples and suppress category imbalance in every batch. Based on the latent dependency between emotions and AUs, an auxiliary branch using graph convolutional layers is added to extract the semantic information from graph topologies. Finally, a re-labeling strategy corrects the ambiguous annotations by comparing their feature similarities with semantic templates. Experiments show that our ULC-AG achieves 89.31\% and 61.57\% accuracy on RAF-DB and AffectNet datasets, respectively, outperform the baseline and state-of-the-art methods.
\end{abstract}


%
\IEEEpeerreviewmaketitle

\section{Introduction}
Facial expression recognition (FER) plays an essential role in realizing human-computer interaction. It can be used in many practical applications, including health monitors, virtual reality, and social robots. Recently, deep neural networks (DNNs) have become the dominant FER methods and achieved excellent performance based on sufficient annotated images and high-speed computing resources \cite{liu2021identity, pham2021facial, zhang2022median}.

\begin{figure}[t]
   \centering
   \includegraphics[width=0.9\columnwidth]{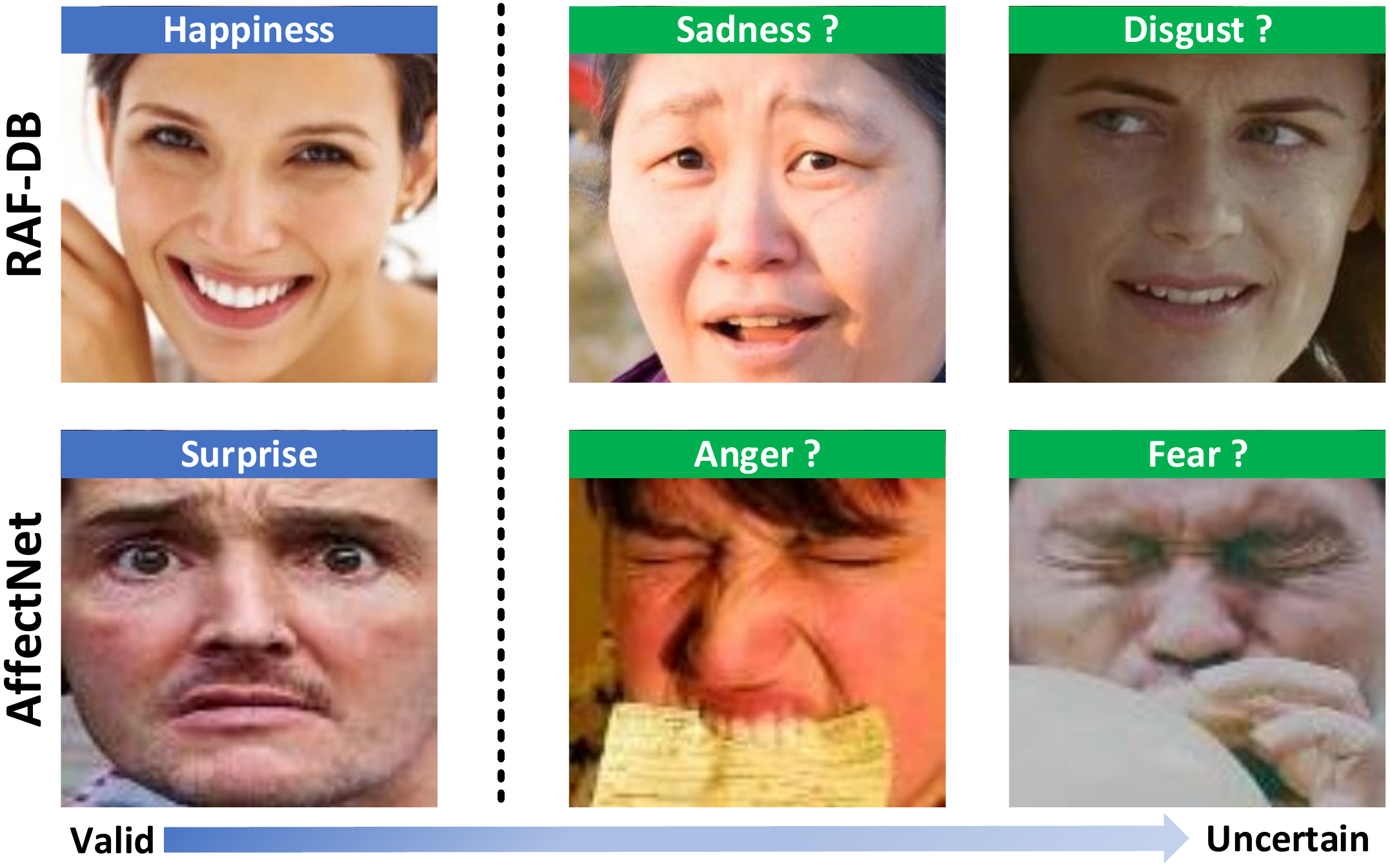}
   \caption{Examples of valid and uncertain images in RAF-DB and AffectNet datasets.}\label{fig:uncertainty}
\end{figure}

Since the training of deep models requires massive data, public FER datasets have been collected in both constrained (e.g., CK+ \cite{lucey2010extended} and Oulu-CASIA \cite{zhao2011facial}) and in-the-wild (e.g., RAF-DB \cite{li2019reliable} and AffectNet \cite{mollahosseini2017affectnet}) conditions. However, for those samples in real-world datasets, their annotations are difficult to maintain consistency in a large-scale manner. As a result, many labels are ambiguous or even incorrect. These may be due to the subjectivity of the annotators and the natural confusion of certain facial expressions. In Fig. \ref{fig:uncertainty}, we show several examples in RAF-DB and AffectNet to illustrate that uncertainty is common in images collected on the Internet. For samples on the left column, annotators can easily make consistent labeling. While for the right column, it is obvious that multiple annotators might have various perspectives on the same sample. In other words, this phenomenon may result in two negative impacts on the model learning: 1) the over-fitting problem will arise due to the considerable proportion of ambiguous samples in the training set; 2) the incorrect labels will mislead the model learning features of specific facial expressions and decrease recognition performance. 

To this end, methods have been studied to mine the knowledge in label space and alleviate the uncertainties. Wang \emph{et al.} \cite{wang2020suppressing} proposed a self-cure network to learn the importance weight of each facial image and suppress uncertain samples by identifying and modifying untruthful labels. Song \emph{et al.} \cite{song2021uncertain} imposed probabilistic masks to capture and weight uncertain samples through an uncertain graph neural network. She \emph{et al.} \cite{she2021dive} exploited auxiliary multi-branch distribution learning, and pairwise uncertainty estimation to solve the ambiguity in both the label space and the instance space. Zhang \emph{et al.} \cite{zhang2021weakly} formulated a noise modeling network based on a weakly-supervised strategy that learned the mapping from feature space to the residuals between clean and noisy labels. 

Recently, the idea of using the relationship among multiple labels has been explored. Chen and Joo \cite{chen2021understanding} incorporated the \emph{triplet} loss into the objective function to embed the dependency between AUs and expression categories.  
Zhang \emph{et al.} \cite{zhang2021facial} designed a unified adversarial learning framework to link the emotion prediction and the joint distribution of dimensional labels. Alternatively, Chen \emph{et al.} \cite{chen2020label} introduced auxiliary label space graphs that cluster samples in neighbor tasks such as landmark detection and AU detection, and leverage the distributions to handle the label inconsistency. Using graphs to solve uncertainty problems was also applied in \cite{zhong2019graph}. Similarly, Cui \emph{et al.} \cite{cui2020label} extracted the dependency between object-level labels and property-level labels, which could be used to revise and generate labels for new datasets. However, these previous methods are still plagued by uncertain samples for the following two reasons: 1) although knowledge including AUs is applied, semantic information is considered from the label level rather than the feature level; 2) the relabeling strategy without constraints is usually rough, which could decrease the reliability of the generated labels. 

In this paper, we perform FER on data with uncertain samples, called \textbf{U}ncertain \textbf{L}abel \textbf{C}orrection via \textbf{A}uxiliary AU \textbf{G}raphs (ULC-AG). ULC-AG consists of two parts: the target branch and the auxiliary branch. For the former, facial features are first extracted through a backbone DNN for every batch of the training data. A weighted regularization module estimates each sample by learning confidence and encourages the model to focus on images with valid labels while considering category imbalance. For the latter, we utilize the same backbone network but change the task to AU detection. It can be regarded as a form of multi-task learning with shared parameters. Then, a graph convolution block is added with all the AU features as nodes to output the semantic feature of each sample. For those images identified as having uncertain labels, we compare their feature similarity with semantic templates and re-label them under the constraint of semantic preserving. Overall, the main contributions can be summarized as follows:

\begin{itemize}
  \item The proposed ULC-AG method mitigates the effects of ambiguous samples and category bias for better facial expression feature learning.
  \item ULC-AG explores semantic information of facial expressions from the auxiliary AU detection task and conducts uncertain label correction under a feature-level constraint.
  \item Our ULC-AG is an end-to-end framework and can achieve superior performance on large-scale FER benchmarks.
\end{itemize}

\section{Proposed Method}
As mentioned above, for public FER datasets, especially with large-scale web images, it is hard to keep all labels high-quality and consistent. To this end, one possible solution is to correct the mislabeled sample with the help of knowledge spaces other than the labels themselves. Inspired by \cite{zhang2021facial,chen2020label}, we introduce the idea of the auxiliary task but focus on the similarity in feature level. The main assumption of this work is that labels of similar samples should have an underlying dependency, which will also be reflected in their feature representation. In this section, we first present an overview of ULC-AG and then elaborate on its crucial modules.

\subsection{Overview of ULC-AG}
\begin{figure}[ht]
   \centering
   \includegraphics[width=0.95\columnwidth]{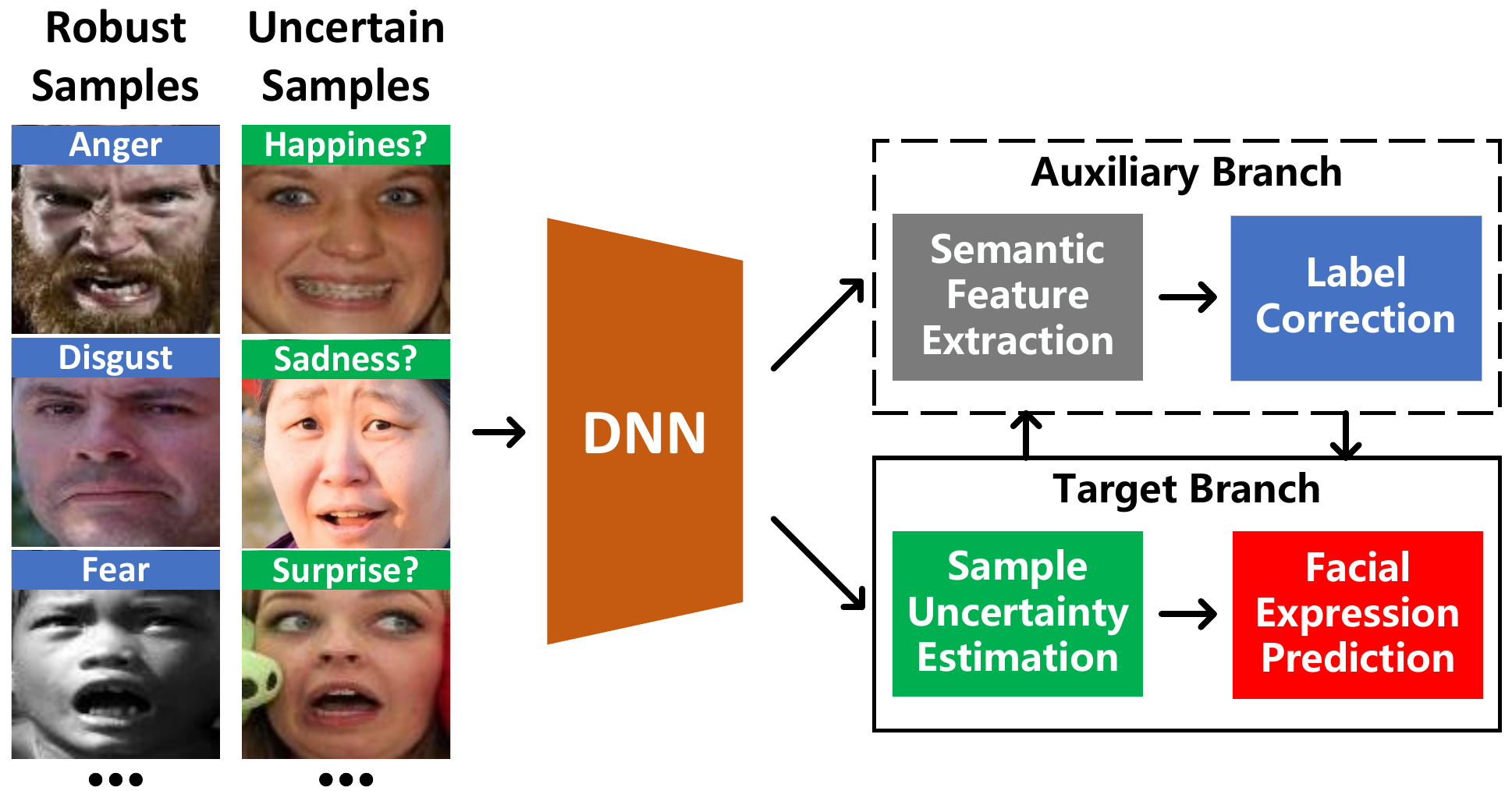}
   \caption{The framework of ULC-AG. It consists of a target branch and an auxiliary branch. The auxiliary branch will not participate in the testing process. Only samples with low confidence labels will be re-annotated.}\label{fig:framework}
\end{figure}

An overview of ULC-AG is illustrated in Fig. \ref{fig:framework}. The ULC-AG contains: 1) a target branch that takes facial features extracted by a pre-trained DNN and computes the annotation confidence using a self-attention layer. These confidence weights will affect the importance of the sample when calculating the classification loss. In addition, the class-oriented weight is computed to deal with category imbalance in the current batch; 2) an auxiliary branch that follows the idea of multi-task learning exploits the same backbone to get AU features of each facial image and feed them into a two-layer graph convolutional network (GCN) \cite{kipf2016semi} for semantic feature extraction. The re-labeling strategy corrects suspicious labels according to the semantic similarity between the low confidence sample and the templates. The whole ULC-AG is an end-to-end framework and the auxiliary branch will not participate in the testing process. 

\subsection{Weighted Regularization}

\begin{figure*}[ht]
  \centering
  \includegraphics[width=1.75\columnwidth]{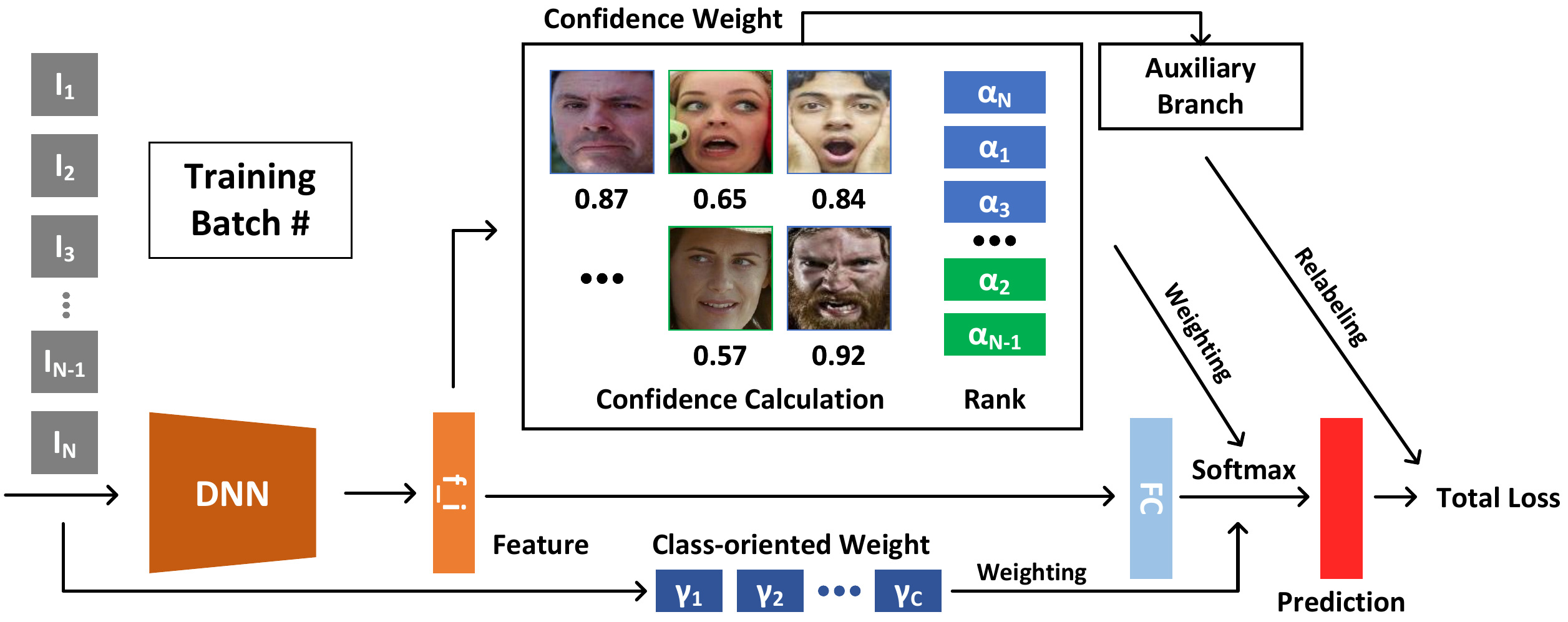}
  \caption{The pipeline of the ULC-AG Target Branch.}\label{fig:target}
\end{figure*}

Fig. \ref{fig:target} illustrates the pipeline of the ULC-AG target branch. To identify the ambiguous samples and estimate their uncertainties, inspired by \cite{wang2020suppressing, hu2019noise}, a self-attention module is employed that consists of a fully connected (FC) layer and the sigmoid function. For a batch of $N$ images, $\bm{F}=[\bm{f}_1,\bm{f}_2,...,\bm{f}_N]\in \mathbb{R}^{D\times N}$ indicates the facial features extracted by the pre-trained DNN, $D$ is the dimension for each facial feature. The confidence weight of the $i$-th sample can be calculated as: 
\begin{equation}
   \alpha_i = Sigmoid(\bm{W}_a^\top\bm{f}_i), 
\end{equation}
where $\bm{W}_a^\top$ denotes the parameters of the self-attention layer. In addition, to prevent the uncertainty caused by category imbalance, we introduce class-oriented weights that are computed as:
\begin{equation}
  \gamma_j = 1 - \frac{N_j}{N}, j\in \{1,2,...,C\}, 
\end{equation}
where $N_j$ is the number of images belonging to class $j$, and $C$ is the number of classes. 

During the model training, it is expected that samples with lower confidence weights should impose less impact, while categories with fewer samples should receive more attention in the current batch. Therefore, we improve the weighted Cross-Entropy (CE) loss proposed in \cite{she2021dive}. Specifically, the loss function for facial expression classifier is formulated as:
\begin{equation}
   L_{wce} = -\frac{1}{N}\sum_{i = 1}^{N}\log\frac{e^{\alpha_i\gamma_{y_i}\bm{W}_{y_i}^\top\bm{f}_i}}{\sum_{j = 1}^{C}e^{\alpha_i\gamma_{y_i}\bm{W}_j^\top\bm{f}_i}},
\end{equation}
where $\bm{W}_j^\top$ denotes the parameters of the $j$-th classifier, $f_i$ is the facial feature, $\alpha_i$ is the confidence weight, $y_i$ and $\gamma_{y_i}$ are the original label and its corresponding class-oriented weight, respectively. According to \cite{liu2017sphereface}, $L_{wce}$ and $\alpha$ are positively correlated. 

After we obtain the confidence weights, the high and low confidence samples in the current batch are divided with a ratio $\varphi$ by using a simple rank regularization approach \cite{wang2020suppressing}, which is formulated as: 
\begin{equation}
   L_{rr} = max(0, \theta - (Avg_h - Avg_l)),
\end{equation}
where $\theta$ is a margin threshold that can be a fixed hyperparameter or updated with training, $Avg_h$ and $Avg_l$ are mean values of confidence weights in high and low groups, respectively.

\subsection{Auxiliary AU Graph Branch}
\begin{figure*}[ht]
  \centering
  \includegraphics[width=1.75\columnwidth]{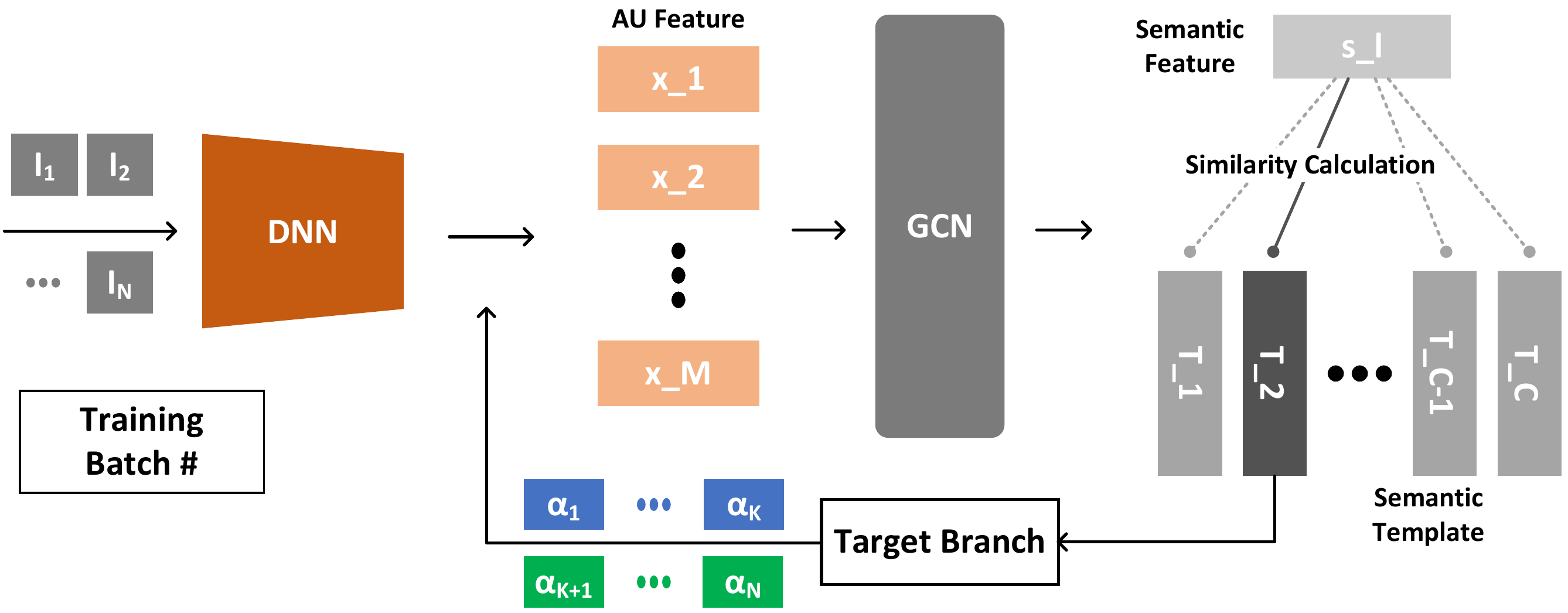}
  \caption{The pipeline of the ULC-AG Auxiliary Branch.}\label{fig:auxiliary}
\end{figure*}

Recent studies reveal that introducing knowledge of the multi-label space can alleviate the effect of ambiguous facial expressions \cite{zhang2021facial,li2021self}. In this work, we choose AU detection as our auxiliary task and construct semantically representative AU graphs because the Facial Action Coding System is an affect description model that has latent mappings with expression categories \cite{friesen1978facial,liu2019facial,liu2021graph}. The AU graph takes individual AU features as graph nodes and the co-occurring AU dependency as graph edges. Fig. \ref{fig:auxiliary} illustrates the pipeline of the ULC-AG auxiliary branch. 

Following the multi-task learning idea, we can get a set of AU features of each image with the same backbone network, $\bm{X}^i=[\bm{x}_1^i,\bm{x}_2^i,...,\bm{x}_M^i]\in\mathbb{R}^{B\times M}$, $B$ and $M$ denote feature dimension and AU number, respectively. Considering the consistency of predefined mappings between expression categories and AUs in large-scale datasets is difficult to guarantee\cite{liu2021sg,lei2021micro}, we exploit a data-driven approach based on the conditional probability to obtain co-occurring AU dependencies from the training set as graph edges, which can be calculated as:
\begin{equation}
	\bm{A}_{p,q} = P(AU_p|AU_q) = \frac{OCC_{p\cap q}}{OCC_q},
\end{equation}
where $OCC_{p\cap q}$ denotes the number of co-occurrences of $AU_p$ and $AU_q$, and $OCC_q$ is the total number of occurrences of $AU_q$. Since the AU co-occurring relationship is actually asymmetry, so $P(AU_p|AU_q) \neq P(AU_q|AU_p)$.

Then, the two are input in a two-layer GCN with each AU as a graph node to extract the semantic feature. Specifically, each graph convolution layer is formulated as: 
\begin{equation}
  \bm{X}'=g(\bm{X},\bm{A})=LeakyRELU(\bar{\bm{A}}\bm{X}\bm{W}_g),
\end{equation}
where $\bar{\bm{A}}$ denotes the normalized $\bm{A}$ with all rows sum to one, $\bm{W}_g$ is the weight matrix to be learned in the current layer.

All the node features outputted by the GCN are fed into a FC layer with sigmoid functions to predict multiple AUs. The binary CE loss is used to train every AU classifier, and the total balanced group loss is defined for the two-layer GCN as:
\begin{equation}
   L_{au} = -\sum_{m = 1}^{M}{\alpha(z_m\log{p_m} + (1-z_m)\log{(1-p_m)})}, 
\end{equation}
where $\alpha$ is the confidence weight, $z_m$ and $p_m$ are the pseudo label and the prediction of $m$-th AU, respectively. The feature $\bm{s}_i\in \mathbb{R}^{1\times M}$ before AU classifiers are treated as the semantic feature of the sample. 

\subsection{Relabeling with Semantic Preserving}
To determine which labels need to be corrected and which new classes should be assigned, we design a semantic preserving strategy (see Fig. \ref{fig:auxiliary}). For the divided two sample sets, a weighted semantic template set $\bm{T}=[\bm{t}_1,\bm{t}_2,...,\bm{t}_C]\in\mathbb{R}^{M\times C}$ for every facial expression category is first generated with the semantic features and the confidence weights of valid samples, which can be formulated as:
\begin{equation}
  \bm{T}_j=\frac{1}{K_j}\sum_{k_j=1}^{K_j}\alpha_{k_j}\bm{s}_{k_j},
\end{equation}
where $K_j$ is the number of the samples with the $j$-th label in the high confidence set. The semantic templates will be dynamically updated throughout the whole training process.

For the ambiguous samples in the low confidence set, we calculate the cosine distance between $\bm{s}_l$ ($l\in \{1,2,...,N-K\}$) and each of $\bm{t}_j$ in the semantic template $\bm{T}$, which can be formulated as:
\begin{equation}
  SP_{s_l,j}=1-\frac{\bm{t}_j \times \bm{s}_l}{\Vert \bm{t}_j\Vert \Vert \bm{s}_l\Vert}, 
\end{equation}
where $\times$ denotes the dot product operation, $K$ is the total number of high confidence samples in each batch. 

Next, for every ambiguous sample, we compare its semantic feature with each semantic template in $T$. The template class with the highest semantic similarity will be assigned to this sample as a new label. Formally, the re-labeling strategy can be defined as: 
\begin{equation}
  y'_i=\begin{cases}
    j, & \mbox{if } SP_{s_l,org}-min(SP_{s_l,j})>0 \\
    y_i, & \mbox{otherwise}  
    \end{cases}
\end{equation}
where $y'_i$ indicates the corrected label, and $j\neq org$, $org$ is the original labeled class.

\subsection{Model Training}
Finally, the total loss function of the whole network can be written as:
\begin{equation}
  L_{total}= \frac{\lambda_1}{2}(L_{wce}+L_{rr})+\lambda_2 L_{au},
\end{equation}
where $\lambda_1$ and $\lambda_2$ are the weighted ramp functions that will change with epoch rounds \cite{laine2016temporal}, which can be computed as follows:
\begin{equation}
	\lambda_1 = \begin{cases}
		\exp(-(1-\frac{epoch}{\beta})^2), & epoch\leq \beta\\
		1, & epoch > \beta
	\end{cases},
\end{equation}
\begin{equation}
	\lambda_2 = \begin{cases}
		1, & epoch \leq \beta \\
		\exp(-(1-\frac{\beta}{epoch})^2), & epoch > \beta
	\end{cases}.
\end{equation}

The weighted ramp functions allow ULC-AG to pay more attention to auxiliary branches in the initial training stage. Since the number of samples accumulated at the beginning is insufficient, it is unable to generate robust semantic features. After a certain number of training rounds, the model will focus more on the target branch to extract discriminative features for final predictions.

\section{Experiments}
\subsection{Datasets}
To evaluate the performance of ULC-AG in tackling label uncertainties, we conduct experiments on two popular FER benchmarks, RAF-DB \cite{li2019reliable} and AffectNet \cite{mollahosseini2017affectnet}. Both datasets have unconstrained conditions and large-scale samples. 

\textbf{RAF-DB} has $15339$ face images with annotations of six basic emotions and neutral. In our experiments, $12271$ and $3368$ samples are used for training and test, respectively.

\textbf{AffectNet} contains close to one million expression images. To ensure a fair comparison, we select samples manually labeled as six basic emotions and neutral for evaluation. The number of images in the training set and the test set is $283,901$ and $3500$, respectively. In addition, automatically labeled samples in AffectNet are used as a set of real noisy data, denoted as \textbf{AffectNet\_Auto}, to verify the ability of ULC-AG in handling uncertain expressions.

Since the AU annotation requires specially trained experts and is time-consuming, it is natural that no AU labels are provided in RAF-DB and AffectNet. To account for this issue, we applied Openface 2.0 \cite{baltrusaitis2018openface} to automatically generate pseudo AU labels, similar to \cite{chen2021understanding,chen2020label}. Note that our ULC-AG utilizes feature-level semantic similarity preserving, which can reduce the negative impact of incorrect pseudo AU labels.

\subsection{Implementation Details}
The ULC-AG is implemented with the Pytorch platform and trained using two Nvidia Volta V100 GPUs. Face images are obtained using MTCNN \cite{zhang2016joint} and further resized to $224\times 224$ pixels as inputs. For the target branch, we choose the ResNet-18 as the backbone DNN which is pre-trained on the MS-Celeb-1M \cite{guo2016ms} dataset as previous methods \cite{wang2020suppressing,she2021dive,wang2020region}. In every iteration, $\varphi$ and $\theta$ are set as $0.8$ and $0.15$, respectively. The initial learning rate is $0.01$, which is updated to $10^{-3}$ and $10^{-4}$ at the $10$-th and $20$-th epoch, respectively. For the auxiliary branch, each GCN layer has 64 channels, and the decayed learning rate is set as 0.005. The auxiliary branch will not participate in the network optimization until $10$ epochs to obtain initial templates. Thus, when the relabeling starts afterward, there is no missing template in the current batch. We choose a batch size $512$ to ensure that every template can be effectively updated during the whole training process. 

\subsection{Ablation Study}
We conduct the ablation experiments to demonstrate the contributions of the proposed modules in this paper. 

\subsubsection{Components evaluation}
ULC-AG aims to solve the influence of uncertain samples during feature learning. The target branch is to suppress low-quality inputs through confidence estimation and weighted regularization, and the auxiliary branch corrects uncertain labels by AU graph construction and semantic similarity constraint, both of which can be flexibly combined with various network architectures. In this experiment, we design four different settings for effectiveness verification. Note that the confidence weight will be calculated but not applied for regularization when only the auxiliary branch works. When the two  branches are not active, ULC-AG is equivalent to a standard ResNet-18. 

As shown in Table \ref{tab:comp}, the independent use of the target branch or the auxiliary branch on the two datasets can significantly enhance the FER performance. In particular, the auxiliary branch makes the greater improvement because it introduces additional semantic information and explicitly manipulates uncertain samples. The best performance is achieved when using the two in collaboration. In other words, ULC-AG can effectively handle the uncertain samples in large-scale data.

\begin{table}
   \centering
   \caption{Evaluation of the components in ULC-AG. '\emph{Target Branch}' applies the weight regularization loss function. '\emph{Auxiliary Branch}' exploits AU graphs and the semantic preserving relabeling.}
   \label{tab:comp}
   \setlength{\tabcolsep}{2.5mm}{
   \begin{tabular}{c|c|c|c}
     \hline
     \textit{Target Branch} & \textit{Auxiliary Branch} & RAF-DB & AffectNet \\
     \hline
     $\times$ & $\times$ & 85.82 & 57.94 \\
     \checkmark & $\times$ & 86.54 & 58.66 \\
     $\times$ & \checkmark & 87.73 & 59.34 \\
     \checkmark & \checkmark & \textbf{89.31} & \textbf{61.57} \\
     \hline
   \end{tabular}}
\end{table}

\subsubsection{Evaluation of data-driven edges}
The data-driven edges introduce important semantic information about AU co-occurring dependencies into the constructed AU graphs. In this experiment, we randomly initialize $A$ to shield edge attributes. Results in Table \ref{tab:edge} show that the data-driven AU co-occurrence matrix can provide AU relationships that approximate the actual distribution, thereby helping the GCN to better extract affective semantic features from the AU graph. 

\begin{table}
	\centering
	\caption{Evaluation of data-driven edges in ULC-AG.}
	\label{tab:edge}
	\setlength{\tabcolsep}{7mm}{
		\begin{tabular}{c|c|c}
			\hline
			Edges & RAF-DB & AffectNet \\
			\hline
			\emph{Random} & 85.06 & 55.79 \\
			\emph{Data-driven} & \textbf{89.31} & \textbf{61.57} \\
			\hline
	\end{tabular}}
\end{table}

\subsection{Evaluation of Handling Uncertain Labels}
To test our re-labeling strategy, we set up comparative experiments under synthetic uncertainty and real uncertainty, respectively. The ResNet-18 baseline and the SCN \cite{wang2020suppressing} also using label repair are selected for comparison.

\subsubsection{Synthetic uncertain samples}
We randomized 10\%, 20\%, and 30\% of the original labels of the training set for RAF-DB and AffectNet, respectively. From Table \ref{tab:syn}, our ULC-AG outperforms another two methods in this experiment. This illustrates the universality of uncertain samples in large-scale facial expression datasets. In addition, as the proportion of uncertain labels increases, the performance degradation of ULC-AG compared to another two methods are also smaller, which further proves the effectiveness of our feature-level semantic similarity constraint.

\begin{table}[h]
   \centering
   \caption{Performance of ULC-AG on datasets with synthetic uncertain samples.}\label{tab:syn}
   \setlength{\tabcolsep}{5mm}{
   \begin{tabular}{c|c|c|c}
     \hline
     Method & Uncertainty & RAF-DB & AffectNet \\
     \hline
     Baseline & 10\% & 80.63 & 57.25 \\
     SCN \cite{wang2020suppressing} & 10\% & 82.18 & 58.58 \\
     ULC-AG & 10\% & \textbf{83.21} & \textbf{59.45} \\
     \hline
     Baseline & 20\% & 78.06 & 56.23 \\
     SCN \cite{wang2020suppressing} & 20\% & 80.10 & 57.25 \\
     ULC-AG & 20\% & \textbf{81.16} & \textbf{58.51} \\
     \hline
     Baseline & 30\% & 75.13 & 52.60 \\
     SCN \cite{wang2020suppressing} & 30\% & 77.46 & 55.05 \\
     ULC-AG & 30\% & \textbf{79.01} & \textbf{56.45} \\

     \hline
   \end{tabular}}
\end{table}

\subsubsection{Real uncertain samples}
Apart from manually annotated samples, we also select AffectNet\_Auto as a training set that has nature uncertain samples for cross-dataset validation, which is rarely considered by previous studies. The automatic labeling algorithm published in the official document has an accuracy of $65\%$ \cite{mollahosseini2017affectnet}. As shown in Table \ref{tab:real}, ULC-AG performs the best when facing real uncertain samples, and the performance growth exceeds that in the synthetic uncertainty experiment. One possible explanation is that the uncertainty in real data is more caused by the insignificant inter-class difference. Our weighted regularization mitigates ambiguities from imbalanced categories, and the semantic information introduced by the auxiliary AU graph further conducts effective label correction.

\begin{table}
	\centering
	\caption{Performance of ULC-AG on datasets with real uncertain samples.}
	\label{tab:real}
	\setlength{\tabcolsep}{7mm}{
	\begin{tabular}{c|c|c}
    \hline
    Method & Uncertainty & AffectNet\_Auto \\
    \hline
    Baseline & Real & 53.23 \\
    SCN \cite{wang2020suppressing} & Real & 55.43 \\
    ULC-AG & Real & \textbf{57.37} \\
    \hline
	\end{tabular}}
\end{table}

\subsection{Visualization}
\subsubsection{Target branch}
\begin{figure}[h]
  \centering
  \includegraphics[width=0.95\columnwidth]{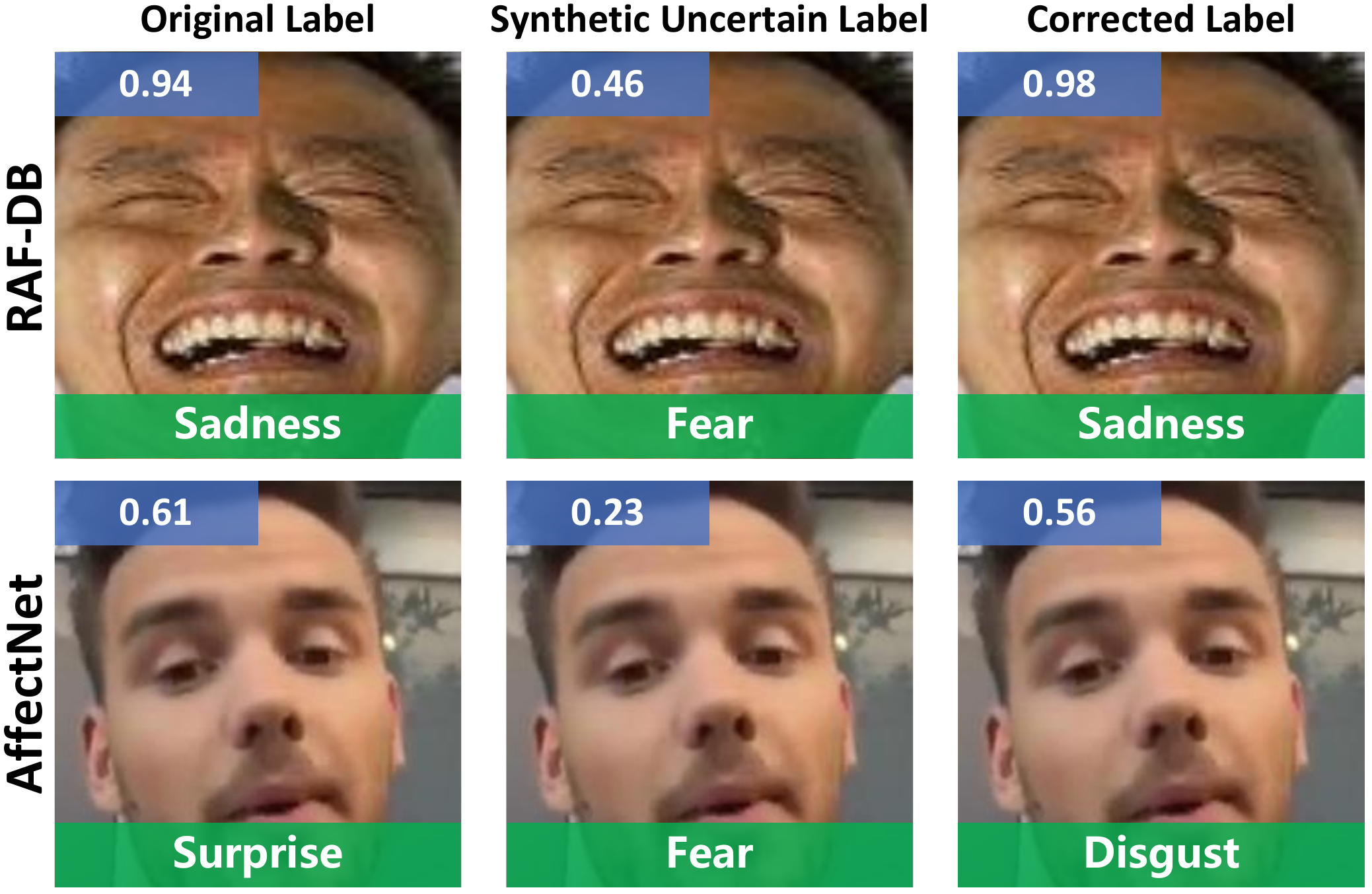}
  \caption{Visualization of target branch. The blue block in the top left corner denotes the confidence value $\alpha$, and the green block at the bottom presents the label of the current sample. Zoom in for better view.}\label{fig:vis-tar}
\end{figure}

Fig. \ref{fig:vis-tar} depicts the visualization of the confidence estimation in the target branch on two examples in RAF-DB and AffectNet datasets. The proposed ULC-AG can successfully perform the label correction on synthetic uncertain labels and adaptively update sample confidence. In particular, in the second line of Fig. \ref{fig:vis-tar}, ULC-AG not only accurately identified the synthetic label but also corrected the originally uncertain sample.

\subsubsection{Auxiliary Branch}
\begin{figure}[h]
  \centering
  \includegraphics[width=0.95\columnwidth]{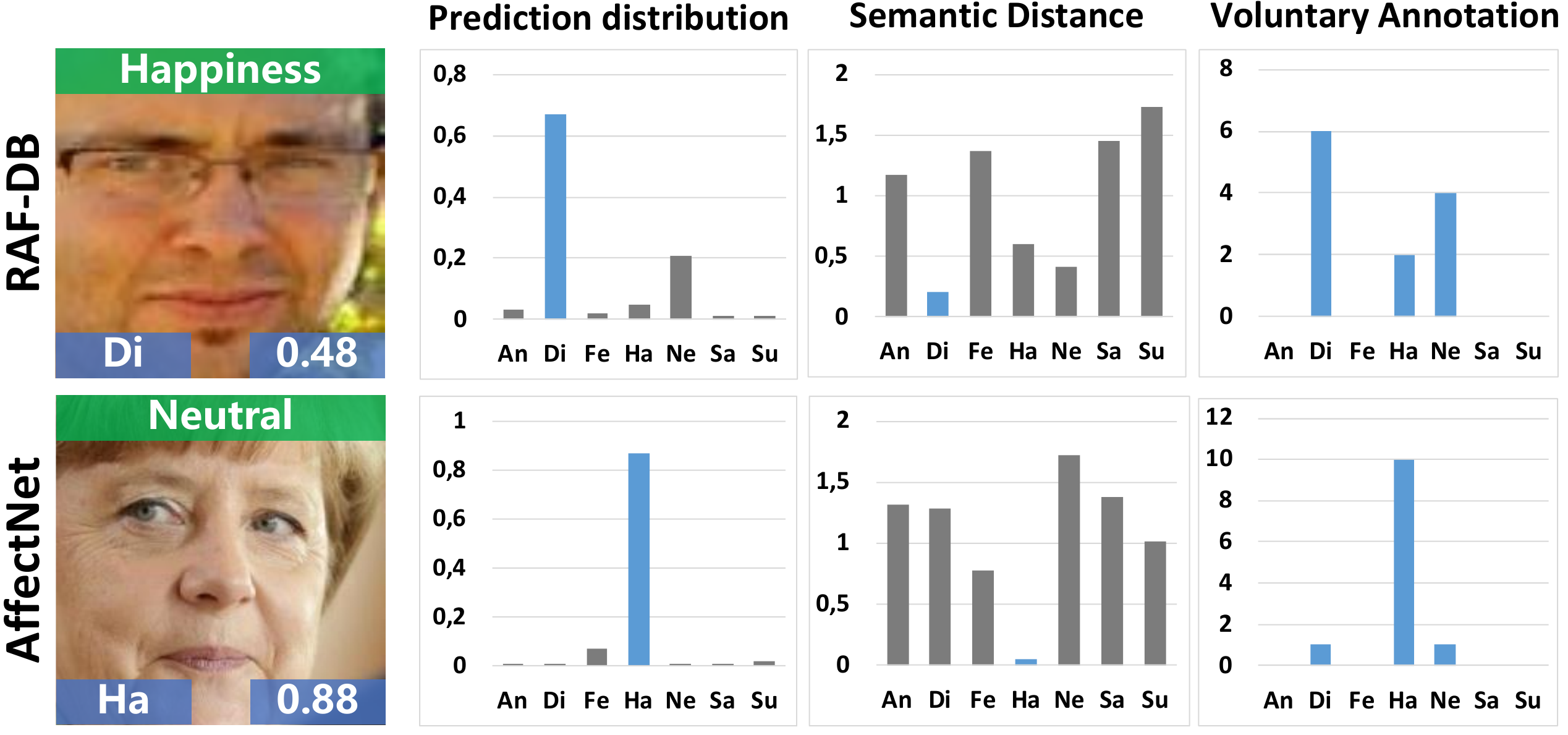}
  \caption{Visualization of auxiliary branch. The green block at the top indicates the synthetic uncertain label, the blue block at the bottom left presents the label after correction, and the blue block at the bottom right is the confidence value $\alpha$ after correction. Zoom in for better view.}\label{fig:vis-aux}
\end{figure}

To further analyze the actual effect of the semantic preserving relabeling in the auxiliary branch, we visualize the key intermediate values in the auxiliary branch on two examples in RAF-DB and AffectNet datasets. In addition, subjective annotations from twelve volunteers are presented to evaluate the label correction strategy. As shown in Fig. \ref{fig:vis-aux}, the semantic feature extracted by ULC-AG can increase the inter-class distance, and the predicted emotion categories are similar in distribution to manual annotations. It reveals that the auxiliary branch can effectively handle uncertain samples to improve the final FER performance.

\subsection{Comparison with the State-of-the-art}
\begin{figure}[h]
	\centering
	\includegraphics[width=1\columnwidth]{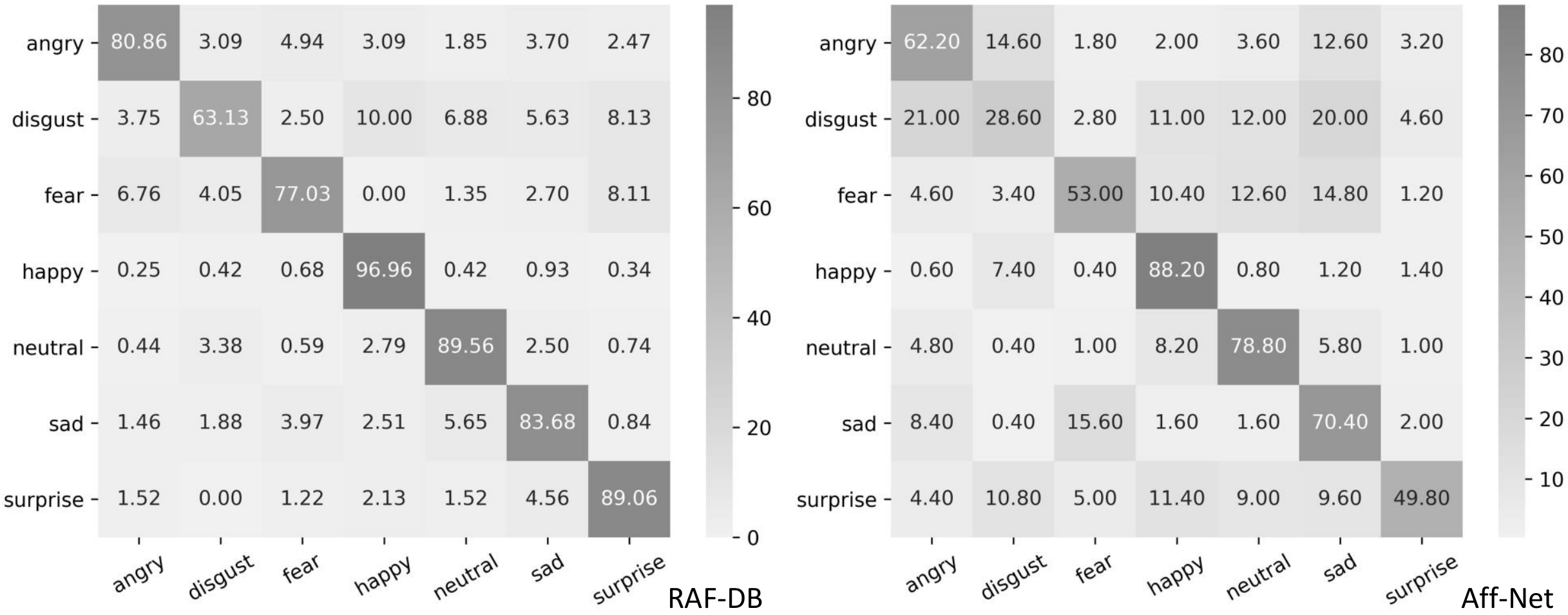}
	\caption{Confusion matrices of ULC-AG. Zoom in for better view.}\label{fig:cm}
\end{figure}

Table \ref{tab:soat} shows the performance comparisons with the state-of-the-art approaches and Fig. \ref{fig:cm} presents the confusion matrices of ULC-AG. To summarize, our method obtains competitive results on both RAF-DB and AffectNet datasets. Although LDL-ALSG \cite{zeng2018facial} and SEIIL \cite{li2021self} introduce different auxiliary tasks to train the network, they only consider the label-level distribution and cannot repair the uncertain samples. In addition, IPA2LT \cite{zeng2018facial}, SCN \cite{wang2020suppressing}, and WSND \cite{zhang2021weakly} explicitly deal with ambiguous images, but the label uncertainties can still mislead feature learning without extra knowledge in side-space and cause performance limitations. Benefiting from the confidence estimation, the data-driven AU graph, and the feature-level constrained label correction, the proposed ULC-AG outperforms all the comparison methods, including NMA \cite{zhang2021facial} that is similar but lacks effective facial representation.

\begin{table}[h]
   \centering
   \caption{Comparisons with the state-of-the-art methods. $\ast$ means RAF-DB and AffectNet are used for training together. $\dagger$ indicates the handling of ambiguous labels is introduced. $\ddagger$ denotes extra knowledge of auxiliary tasks is considered.}\label{tab:soat}
   \setlength{\tabcolsep}{4.5mm}{
   \begin{tabular}{c|c|c|c}
     \hline
     Method & Year & RAF-DB & AffectNet \\
     \hline
     IPA2LT$^\dagger $\cite{zeng2018facial} & 2018 & 86.77 & 55.11$^\ast$ \\
     SCN$^\dagger $ \cite{wang2020suppressing} & 2020 & 88.14$^\ast$ & 60.23\\
     RAN \cite{wang2020region} & 2020 & 86.90 & 59.50 \\
     LDL-ALSG$^\ddagger$ \cite{chen2020label} & 2020 & 85.53 & 59.35 \\
     SPWFA-SE \cite{li2020facial} & 2020 & 86.31 & 59.23 \\
     SEIIL$\ddagger$ \cite{li2021self} & 2021 & 88.23 & / \\
     WSND$^\dagger$ \cite{zhang2021weakly} & 2021 & 88.89 & 60.04 \\
     IDFL \cite{li2021learning} & 2021 & 86.96 & 59.20 \\
     NMA$^{\dagger\ddagger}$ \cite{zhang2021facial} & 2021 & 76.10 & 46.08 \\
     \hline
     ULC-AG$^{\dagger\ddagger} $ & Ours & \textbf{89.31} & \textbf{61.57} \\
     \hline
   \end{tabular}}
\end{table}

\section{Conclusion}
In this paper, we proposed the ULC-AG framework to alleviate the uncertainties in facial expression images. The weighted regularization helped the model identify ambiguous images and balance categories. The relabeling strategy with semantic preserving corrected the suspicious labels through the auxiliary AU graph. Experiments on two large-scale datasets showed that ULC-AG achieved superior results and was robust to uncertain labels. In the future, other auxiliary tasks such as landmark detection and intensity estimation can be considered, and ULC-AG can be extended to generate annotations for unlabeled data and make multi-task predictions.

\newpage

\balance
\bibliographystyle{IEEEtran}
\bibliography{IEEEabrv, ref}

\end{document}